# Automatic Detection of Neurons in NeuN-stained Histological Images of Human Brain

Andrija Štajduhar[1,2,3,*], Domagoj Džaja[1], Miloš Judaš[1], Sven Lončarić[2]

[1] Croatian Institute for Brain Research, School of Medicine, University of Zagreb, Šalata 3, Zagreb, Croatia
[2] Faculty of Electrical Engineering and Computing, University of Zagreb, Unska 3, Zagreb, Croatia
[3] Montreal Neurological Institute, McGill University, 3801 University Street, Montreal, Quebec, Canada

*Abstract*—In this paper, we present a novel use of an anisotropic diffusion model for automatic detection of neurons in histological sections of the adult human brain cortex. We use a partial differential equation model to process high resolution images to acquire locations of neuronal bodies. We also present a novel approach in model training and evaluation that considers variability among the human experts, addressing the issue of existence and correctness of the golden standard for neuron and cell counting, used in most of relevant papers. Our method, trained on dataset manually labeled by three experts, has correctly distinguished over 95% of neuron bodies in test data, doing so in time much shorter than other comparable methods.

*Index Terms*—Neuron detection, Partial differential equations, Brain histology, NeuN

## 1 INTRODUCTION

SINCE the establishment of cell theory in early 19th century, microscopy has had an important role in life sciences. Scientists made many discoveries observing and studying the cells, fundamental building units of life. In this paper, we are interested in a specific type of cells - the neurons. More precisely, we focused on finding and quantifying the locations of neurons in histological sections of human brain tissue. A historical overview of segmentation methods of different cell types can be found in [1].

### 1.1 Neuron Quantification

Many neurological and psychiatric diseases cause changes in the number of neurons in the brain. These changes are often subtle and can only be proven by quantification [2–4]. Currently, neuron counting is done manually during which only a small sample of neurons is marked and counted, and the total number is estimated with large errors. This approach has several shortcomings. The process of manual counting is tedious, repetitive and uses a vast amount of time. It requires expertise in the field of neuroanatomy and stereology and a skilled researcher may mark up to 15 neurons per minute. The fact that there are more than $10^5$ neurons in an average section makes the process of precise counting of each neuron infeasible. Also, researchers doing the counting may have strong biases in recognition of neurons which results in inter- but also intraobserver variability, putting in question validity of the number of counted neurons. Statistical methods for approximation of total number of neurons base the projections on small manual counts which may yield large errors.

Automatic detection of neurons in histological sections would allow for an objective classification on a large scale, free of human bias, and provide an equivalent analysis on all available preparations. Precise and fast neuron identification would enable researchers to gain better insight in neuron organization and distribution, small network and column formation, delineation of brain area boundaries, and comparison of various pathologies. It would thus significantly improve and speed up quantitative studies.

### 1.2 Histological Imaging

Histological staining is a method used to examine cellular and structural layout of tissue. In this technique, the tissue of interest is preserved using chemical fixatives and sectioned, i.e. cut into very thin sections. Uncolored, these sections have very little variation in colors/shades, so they are treated with various stains to increase the contrast in the tissue.

Some of the most commonly used methods for staining are Nissl method, a classical staining for the cells in brain tissue, and NeuN, an immunohistochemical method which indicates neuronal cell bodies in histological preparations.

Both techniques have their advantages and drawbacks. Nissl method is more affordable but stains all cells in the brain tissue, while NeuN method only stains neurons. In our considerations, we decided that the case in which only neurons are visible is more appropriate for neuron quantification. After preparation and slicing of the tissue, one may choose to stain only several sections with NeuN and use them for neuron quantification and have the other sections stained with more classical staining for other purposes. A work that shows comparison of the two methods is presented in [5].

This publication was co-financed by the European Union through the European Regional Development Fund, Operational Programme Competitiveness and Cohesion, grant agreement No. KK.01.1.1.01.0007, CoRE - Neuro and the Canada First Research Excellence Fund, awarded to McGill University for the Healthy Brains for Healthy Lives initiative.

*Andrija Štajduhar (email: andrija.stajduhar@hiim.hr, corresponding author), Domagoj Džaja (email: domagoj.dzaja@hiim.hr) and Miloš Judaš (email: milos.judas@hiim.hr) are also with Center of Research Excellence for Basic, Clinical and Translational Neuroscience, Šalata 12, Zagreb, Croatia.

Sven Lončarić (email: sven.loncaric@fer.hr) is also with Centre of Research Excellence for Data Science and Advanced Cooperative Systems, Unska 3, Zagreb, Croatia.

## 1.3 Related Work

Although methods for automatic identification and counting of non-neuronal cells might be successful and in widespread use [1], they are usually based on very basic concepts which are appropriate for detection of regularly shaped (mostly oval-like) objects with distinguishable background, like in [6]. However, cells in brain tissue may form complex, irregular formations where cells are located close to one another – these can easily be miscounted as one cell even by a human investigator, which is probably the reason for a lack of accuracy and speed in current automated methods for this task.

It is remarkable that, to the best knowledge of the authors, the first paper published with the focus on automatic neuron segmentation [7] appeared in 2008, and was followed by two papers [8, 9] a year later. We mention and provide a brief outline of some of these related methods.

In [7], the authors introduced an algorithm whose purpose is to obtain coordinates of individual neurons in digitized images of Nissl-stained preparations of the cerebral cortex of the Rhesus monkey. Their approach combined image segmentation and machine learning methods, namely active contour segmentation seeded with use of watershed method for detection of outlines of potential neuron cell bodies and a multilayer perceptron (MLP) containing a single, 4-node hidden layer. The MLP was used to distinguish between neurons and non-neurons based on features derived from the segmentation, such as segment area, optical density, contour gyration and similar. The authors state that their method positively identifies $86 \pm 5\%$ neurons with $15 \pm 8\%$ error (mean $\pm$ SD) on a range of Nissl-stained images at 10x magnification.

In the second article [8], the authors propose a novel multi-layer shape analysis of blobs that are candidates for being identified as neurons. Although not significantly improving in accuracy ($87 \pm 6\%$ positively identified neurons) over the first method, the main segmentation idea is multilevel thresholding and analysis of the obtained level structure for separating neurons that are closely located. Some shape descriptors for distinguishing neuron bodies from other structures are also used.

We can observe that these methods are not very reliable in terms of accuracy. Probably the biggest challenges are closely located or overlapping neurons that are often identified as a single neuron cell, and shapes too different from predefined oval-like contours that are not being identified as neurons. Present noise and speckles may also cause problems for deformable models [10]. Besides inaccuracy being the obvious shortcoming, the methods suffer from high computational cost - as authors state, it takes days to process a single, medium size section.

Authors in [9] report over 90% detection accuracy on NeuN images of rat cortex using three-step image processing pipeline that includes significant image pre-processing, morphological filtering and finally model-based filtering, with many sub-steps. The last two steps are essential for splitting of closely located neurons, indicating the importance of overcoming this issue in neuron detection. Although more precise than the first two methods, this approach is relatively more computationally expensive - its analysis time is about one fourth of that of the manual effort, thus still taking days or more than a week, depending on a section size, to process a whole section, with some manual work still included.

## 2 MATERIALS AND METHODS

Our aim was to develop a procedure for automatic identification of all neurons visible in the tissue and distinguish neurons from noise and artifacts, while doing so in significantly shorter time than when done by humans.

For the method development, we used histological sections of the adult human prefrontal cortex, stained with NeuN immunohistochemistry method, from the Zagreb Brain Collection [11]. Preparations were digitized using Hamamatsu Nanozoomer 2.0 scanner using a single optical plane at 40x magnification, corresponding to 0.226µm/pixel resolution. Because of the size and the special image format of scans produced by the scanner (.ndpi), we exported smaller TIFF images and processed them separately. In our experiments we used 10µm thick sections. We concluded that images of a slightly lower in-plane resolution also yielded satisfying results. Therefore, we subsampled the original scans by a factor of 2, and used the images at 0.452µm/pixel resolution. This enabled greater processing speed and lower memory load. Further subsampling would result in slight loss of the method's performance. An example portion of a histological image is shown in Fig. 1.

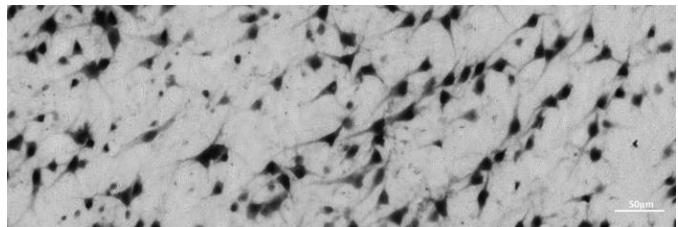

*Figure 1. Example of NeuN-stained histological section image with visible neurons. Image is obtained under 40x magnification which corresponds to 0.226µm/pixel resolution.*

As shown in the Results section, it is not easy to measure efficiency of the developed method because there is no absolute mutual agreement between the experts on what is to be identified as neuron. Some neuron bodies that have low color intensity are considered to be in the tissue of adjacent section, and are not to be counted among the neurons of current section. There is also disagreement between the researchers doing manual counting on some neurons of smaller body size.

Therefore, our goal is to develop a model that agrees with the human experts in the similar fashion as they agree between themselves, i.e. does not differ in neuron identification from human experts more than they mutually differ.

In this task, we are faced with two main challenges - recognize neuron bodies and distinguish between closely located ones, despite artifacts and noise. One could do this by looking for local minima on the image and identify each that is darker than some predetermined intensity as a neuron center. This indeed very basic idea is fundamental in our approach, especially given the fact that the neurons are most commonly darker in their center due to the dye uptake during staining. However, a single neuron may have many local image intensity minima (noise) that would in this way be identified as multiple neurons. One could process the located minima in a way that those located closely to each other may be combined into a single one. Unfortunately, this approach would often merge some small neurons; the distance between local minima that may appear in a single neuron might be greater than a few smaller closely located neurons combined.

Although these extra minima are not present due to the noise added during the image acquisition, but are rather inherent by the nature of neuron representation on the stained preparation, we steered our approach inspired by denoising techniques in image processing. Visualizing images of neurons as functions of two variables in 3D (see Fig.2), the third dimension being image gray intensity, it seems that cone-like structures of neuronal bodies should be *smoothed* in some way. For that purpose, we decided to approach the problem of finding neuron cells from a new perspective, one that is not usually seen in literature on this subject, but that seems more natural and is fit for the specific purpose of detecting neurons on a histological preparation. The idea is to capture the diffusive nature of this staining, i.e. dye uptake in the neurons which occurs mostly in the neuron center, since here, NeuN is a neuronal antigen that binds mostly in the neuron nuclei. Therefore, we investigated methods that are inspired by diffusion process and are described by partial differential equations that govern such processes in nature. Some of the methods we considered were shock filtering [12], $L_0$ gradient minimization [13], global image smoothing based on weighted least squares [14], bilateral filtering [15]. Our approach avoids exhaustive preprocessing, complicated pipelines and multiple steps in tissue image processing, such as compensation for artifacts and speckles, noise reduction, background removal and similar, commonly seen in various approaches for cell detection [6–8].

*2.1 PDE models*

Partial differential equations (PDEs) introduce a new approach to digital image processing. Extensive mathematical results relying on strong theoretical foundations are available and provide stable numerical schemes. Some *ad hoc* filters that were developed in image processing were later justified by PDE theory [16,17], which continued to produce new, more efficient filters for various purposes based on studied mathematical properties of these filters.

One well known denoising technique is filtering with a Gaussian kernel with purpose to remove noise and smooth the image. As it can be shown, it is equivalent to applying the heat equation

$$u_t = \Delta u \qquad (1)$$

to the image, in terms of its discrete, two-dimensional domain with initial condition being $u_0 = f$, with $f$ being the original image. Our focus was to exploit properties of PDE based models to *smooth* the extra minima that are present in the neuronal bodies. In the two-dimensional case of processing an image, we use the initial problem

$$\begin{cases} u_t = u_{xx} + u_{yy}, & \Omega \times \langle 0, T \rangle \\ u = I, & \Omega \times \{t = 0\} \end{cases} \qquad (2)$$

with $I = I(x, y)$ representing the image intensity we consider and $\Omega$ being the image domain. We assume Dirichlet boundary condition as we do in our considerations of PDE application on images.

However, this model does not yield the desired results. The image intensities are diffused heavily before the extra minima are removed and during the process neurons with lighter intensities are fused with the image background. This directs us to modify our approach in a way that we maintain the smoothing effect but preserve contours of neuronal bodies. We wish diffusion to have effect in the regions of neuronal bodies but not to fuse intensities over the neuron edges. Anisotropic diffusion provided satisfying results.

*2.2 Anisotropic diffusion filtering*

We write Eq. 1 in its divergence form

$$u_t = \Delta u = \text{div}(\nabla u)$$

and insert a diffusion control function inside the divergence, as authors did for denoising purpose in [18]. The function exploits the fact that the image gradient is large near the edges, so is provided with the gradient as a function argument. We obtain

$$\begin{cases} u_t = \text{div}(g(|\nabla u|^2)\nabla u), & \Omega \times \langle 0, T \rangle \\ u = I, & \Omega \times \{t = 0\} \end{cases} \qquad (3)$$

known as the Perona-Malik model with usual choice of $g$ as a decreasing function that inhibits the diffusion effect in image areas with large gradient.

$$g(s^2) = \left(1 + \frac{s^2}{\lambda^2}\right)^{-1}$$

Here, $\lambda$ is a diffusion scaling parameter that will be defined during the method optimization. This equation has been extensively studied (see, for example [19] and references therein), and important theoretical results are available.

In Eq. 3, the diffusivity control function $g$ introduces nonlinearity. However, with its values being between 0 and 1, it does not introduce additional restrictions on the choice of time step $\Delta t$ for numerical stability in the explicit finite differences discretization scheme we used. Taking $\Delta x = \Delta y = h$ for discrete derivatives, it can be shown that for an 8-neighbour scheme, maximum time integration constant is $\Delta t/h^2 = 1/7$ [20].

As for the total time of the integration $T$, we chose 12 iterations in the discretization process, as described in Results section, resulting in $T = N \times \Delta t = 12 \times 1/7 \times h^2 \approx 1.71h^2$. By letting the scaling factor $\lambda$ to infinity, we obtain $g = 1$, thus removing the nonlinearity and reducing the Eq. 3 to its linear form (2). We can now relate $T$ to the full width at half maximum (fwhm) for linear case. Using

$$\text{fwhm} = 4\sqrt{T \ln 2}, \quad (4)$$

we obtain fwhm $= 4.36h$, which is about one fifth of the diameter of an average neuron, and almost half of the diameter of a smallest interneurons in our images. To use the method on images of another resolution, one should account for the difference in pixel size to preserve the same diffusion properties. Using the Eq. 4, we establish relation between number of iteration steps needed in processing of images of different resolution. For a constant fwhm, from Eq. 4 one obtains

$$h_1^2 N_1 = h_2^2 N_2,$$

$h_i$ being the distance between integration nodes and $N_i$ being the number of iterations. For example, in images of half the resolution, one should perform one fourth of the number of iterations.

The Eq. 3 is forward-backward parabolic type and one could expect some instabilities. However, practical implementations of the process work satisfactorily. Essentially, the only instability observed in numerical schemes is the *stair-casing* effect, where a smooth step edge develops into piecewise linear segments separated by jumps. More details on this can be found in [19]. Fortunately, this does not influence local minima, so is of no concern to us. The effect of applying the diffusion process on a portion the image containing neurons is shown in Fig. 2.

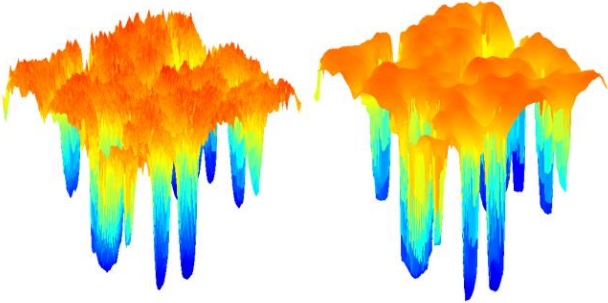

*Figure 2: Result of processing a histological image patch using Perona-Malik model. On the left-hand side is the original data with visible large local signal variation. The data shown on the right-hand side is smoother with less local variation while the intensity minima in neuron nuclei are preserved.*

It was shown in [18] that one of the properties of this equation is non-introduction of new extrema, i.e. if a point in the processed image is a local minimum at some point in time, it was previously a local minimum in the original image as well. Nevertheless, the equation will eliminate minima that do not differ much in pixel value from pixels in their surrounding or are global in some small vicinity in short time. It will keep minima with larger vicinity for a longer time, which is in our case exactly the desired property. The implementation and application of the model on the image was done in MATLAB.

*2.3 Method outline*

We use the model to acquire a single local minimum per neuronal body. However, some minima remain in the image outside the neuronal bodies so from the PDE model we obtain *candidates* for neuron locations, some of which must be ruled out by other means.

We first perform selection of candidate points based on their grayscale intensity, level of which was derived from the data that was manually labeled by human experts. If a candidate point is brighter than a predefined threshold, it means that it is a point outside a neuron, but is rather being part of the background. Another example are image regions that are a part of a neuron whose main body is located in the next section and is not to be counted among neurons in the current one.

Since image may contain noise and artifacts, small but dark objects are possible to appear. To detect minima that are in such areas we performed image thresholding at above mentioned intensity level. A local minimum that is in a blob whose area is smaller than the minimum neuron size is also excluded.

There are four parameters used in this method – gray intensity threshold and minimum cell size were derived from measurements made on manually labeled data. Diffusion scaling constant ($\lambda = 11$, for $h = 1$ and image grayscale range $[0,255]$) in function $g$ and number of iterations ($N = 12$) in numerical scheme were obtained by optimization described in the Results section.

## 3 RESULTS

As mentioned earlier, it is not always clear whether an object in the image should be considered a neuron or not, which is reflected by the presence of variability in manual labeling between the experts. We measured the agreement between two manual labeling results on the same data by dividing the number of neurons in the image that both raters labeled as neurons with number of neurons labeled in total by any rater. In other words, we divided the number of neurons in the intersection with number of neurons in the union of the two raters' labeling output,

$$\delta(i,j) := \frac{|i \cap j|}{|i \cup j|} \quad (5)$$

The agreement was measured by analysis of dataset that was independently manually labeled by three experts. This dataset consisted of 10 images containing over 550 neurons. Those identified by all three experts made 80.33%, identified by two made 8.88%, and those identified by only a single expert made 10.80% of total number of neurons. This results in average pairwise mutual agreement between the raters of $86.88 \pm 0.77\%$.

Taking this into account, and that there is no true baseline that we can rely on, we sought to reproduce the work of raters, which meant creating a method whose performance would be indistinguishable from that of humans.

| Method | Brain tissue type | Image resolution | Accuracy | Execution time |
| --- | --- | --- | --- | --- |
| Inglis et al [8] | Rhesus monkey cortex | 10x, 1.5 µm/px | 86 ± 5% | Several days |
| Sciarraba et al [9] | Human cortex | 40x, 0.26 µm/px | 87 ± 6% | 2 Days |
| Oberleander et al [10] | Rat cortex | 40x, 0.26 µm/px | >90% | Several days |
| Our method | Human cortex | 40x, 0.26 µm/px | >95% | 30 minutes |

*Table 1: Comparison with other relevant methods. We can observe that our method has improved accuracy, while the execution time is significantly shorter.*

It should nevertheless be consistent in terms of giving the same result for the same input and much faster than humans. In our experiments we also noticed the experts' high inconsistency on the same dataset. In an experiment with repeated data, an expert achieved agreement as low as 81.41%, indicating high intraobserver variability.

The reason we did not choose various Kappa statistics is non-existence of true negatives. Every image element that was not labeled as a neuron by any expert would represent an agreement between the raters, leading to an artificially large agreement on true negatives.

In diffusion parameter optimization (diffusion scaling constant $\lambda$ and number of iterations $N$) for the presented method, a ratio of average agreement between the experts and average agreement of experts with the method measured how similar the method's performance is to humans',

$$\Delta(E; m) := \frac{\sum_{i,j \in E, i \neq j} \delta(i,j)}{\sum_{i \in E} \delta(i,m)} \quad (6)$$

where $E$ represents experts and $m$ represents the method output. The best obtained value was $\Delta=0.9784$, which indicates a high similarity in the method's performance with human performance. It is however much faster than humans, and faster than methods found in literature. Total runtime for executing MATLAB code and processing a whole section containing approximately 600,000 neurons was about 20 minutes on an average workstation. Considering the total number of neurons counted on all images, the average number reported by three experts is $547.33 \pm 11.14$, while the method counted 562 neurons which is about 15 neurons or 2.68% more. Table 1 shows comparison of our method with other relevant methods for automated identification of neurons.

After creating the method and obtaining the parameters in described fashion, we chose another set of images totaling approx. 660 neurons on which we used our method for automatic detection of neurons. The results were validated by an expert and the method achieved high accuracy (0.9541), sensitivity (0.9674), specificity (0.9858) and F1-score (0.9765).

*3.1 Application on BigBrain data*

After development and evaluation, we applied our method on the data from the BigBrain project [22], a freely accessible high-resolution 3D digital atlas of the human brain. These sections were cut at 20µm thickness and stained with Merker cell-body stain. To detect all visible cells, we used the method on sections from primary visual cortex digitized at 1µm/px resolution. Although precise evaluation of the cell detection accuracy is not feasible, visual inspection showed satisfying results. After the procedure, we calculated a density map of identified cells, which is shown on the right-hand side of the Fig. 3. Each pixel on the map represents number of cells found in a single frame of the mesh overlaid on the original image. Size of the mesh frames was 27x27 pixels, which corresponds to an area of 729µm$^2$. Resulting image was also slightly blurred to account for granularity and for better visual perception. The obtained cell density map clearly reveals cortical layers and points to correspondence between cell density and optical image intensity.

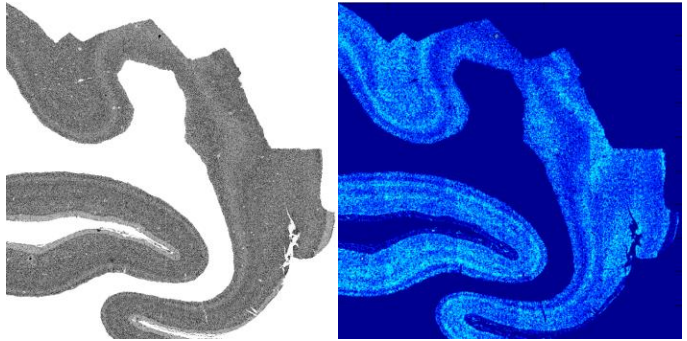

*Figure 3. Left: Image of manually segmented gray matter of primary visual cortex from the BigBrain data. Right: Cell density map obtained using the developed method. Visually darker regions in the histological image coincide with cell density.*

4 CONCLUSION

In this paper, we have presented a novel use of anisotropic diffusion process for processing and analysis of histological images and developed a fast and accurate method for neuron detection. We have also demonstrated the use of the developed method by creating a cell density map of a whole slice of human visual cortex. Given the locations of neurons one could also develop measurements of density, distinguish between different neuron populations by parameters of their somata and observe distribution, size and shape of neurons and obtain other local tissue features. During our experiments, we measured the extent of disagreement between raters on the same dataset and inconsistency within a single rater on repeated data. By this, we addressed the issue of existence and correctness of the golden standard for neuron and cell counting, used in most of the relevant papers.

*Future Work*

A natural extension to considering the inconsistency between the raters would be to introduce a fuzzy-type neuron identification. As some cells are cut in two during slicing, they will appear with different grayscale intensity. Cell fractions between 0 and 1 could be counted. The method is further applicable in developing experiments that include neuron classification and distribution measuring, identification of cortical layers and measuring of cortical thickness.

ACKNOWLEDGMENT

Authors extend their gratitude to Dora Sedmak and Goran Sedmak from Croatian Institute for Brain Research (CIBR), School of Medicine, University of Zagreb, Zagreb, Croatia, for their effort in neuron labeling and helpful discussions. Special thanks to Claude Lepage from Montreal Neurological Institute (MNI), McGill University, Montreal, Canada, for reading the paper thoroughly and providing constructive feedback.

REFERENCES

[1] E. Meijering, "Cell Segmentation: 50 years down the road," *IEEE Signal Processsing Magazine,* vol. 29, no. 5, pp. 140-145, 2012.

[2] D. P. Pelvig, H. Pakkenberg, A. K. Stark and B. Pakkenberg, "Neocortical glial cell numbers in human brains," *Neurobiology of aging,* vol. 29(11), pp. 1754-1762, 2008.

[3] B. Pakkenberg, H. J. G. Gundersen, "Total number of neurons and glial cells in human brain nuclei estimated by the disector and the fractionator," *Journal of microscopy,* vol. 150(1), pp. 1-20, 1988.

[4] B. Pakkenberg, "Total nerve cell number in neocortex in chronic schizophrenics and controls estimated using optical disectors," *Biological psychiatry,* vol. 34(11), pp. 768-772, 1993.

[5] R. Gittins, P. J. Harrison, "Neuronal density, size and shape in the human anterior cingulate cortex: a comparison of Nissl and NeuN staining," *Brain research bulletin*, vol 63.2, pp. 155-160, 2004.

[6] T. Liu, G. Li, J. Nie, A. Tarokh, X. Zhou, L. Guo, S. T. Wong, "An automated method for cell detection in zebrafish, " *Neuroinformatics*, vol. 6(1), pp. 5-21, 2008.

[7] M. Sciarabba, G. Serrao, D. Bauer, F. Arnaboldi and N. A. Borghese, "Automatic detection of neurons in large cortical slices," *Journal of neuroscience methods,* vol. 182(1), pp. 123-140, 2009.

[8] A. Inglis, L. Cruz, D. L. Roe, H. E. Stanley, D. L. Rosene and B. Urbanc, "Automated identification of neurons and their locations," *Journal of microscopy,* vol. 230(3), pp. 339-352, 2008.

[9] M. Oberlainder, V. J. Dercksen, R. Egger, M. Gensel, B. Sakmann, H. C. Hege, "Automated three-dimensional detection and counting of neuron somata," *Journal of neuroscience methods*, vol. 180.1, pp. 147-160, 2009.

[10] T. McInerney and D. Terzopoulos, "Deformable models in medical image analysis: a survey," *Medical image analysis,* vol. 1(2), pp. 91-108, 1996.

[11] M. Judaš, G. Šimić, Z. Petanjek, N. Jovanov-Milošević, M. Pletikos, L. Vasung, M. Vukšić, and I. Kostović, "The Zagreb Collection of human brains: a unique, versatile, but underexploited resource for the neuroscience community," *Annals of the New York Academy of Sciences* vol. 1225.S1, pp. 105-130, 2011.

[12] G. Gilboa, N. Sochen, and Y. Zeevi, "Regularized shock filters and complex diffusion." *Computer Vision—ECCV* 2002: 399-413. 2002.

[13] L. Xu, C. Lu, Y. Xu and J. Jia, "Image Smoothing via $L0$ Gradient Minimization", *ACM Transactions on Graphics*, Vol. 30, No. 5 (*SIGGRAPH Asia 2011*), 2011.

[14] D. Min, S. Choi, J. Lu, B. Ham, K. Sohn, and M. N. Do, "Fast Global Image Smoothing Based on Weighted Least Squares", *IEEE Trans. on Image Processing*, vol. 23.12, pp. 5638-5653, 2014.

[15] S. Paris and F. Durand, "A Fast Approximation of the Bilateral Filter using a Signal Processing Approach", MIT technical report, *International journal of computer vision,* vol. 81.1, pp. 24-52, 2009.

[16] O. R. Vincent and O. Folorunso, "A descriptive algorithm for Sobel image edge detection," *Proceedings of Informing Science & IT Education Conference (InSITE),* vol. 40, pp. 97-107, 2009.

[17] L. Ding and A. Goshtasby, "On the Canny edge detector," *Pattern Recognition,* vol. 34(3), pp. 721-725, 2001.

[18] P. Perona and J. Malik, "Scale-space and edge detection using anisotropic diffusion," IEEE Transactions on pattern analysis and machine intelligence, vol. 12(7), pp. 629-639, 1990.

[19] P. Guidotti, "Anisotropic Diffusions of Image Processing from Perona-Malik on," *Advanced Studies in Pure Mathematics,* vol. 99, pp. 1-30, 2014.

[20] G. Gerig, O. Kubler, R. Kikinis, F. A. Jolesz, "Nonlinear anisotropic filtering of MRI data", *IEEE Transactions on medical imaging*, vol. 11(2), pp. 221-232. 1992.

[21] S. Kichenassamy, "The Perona-Malik paradox." *SIAM Journal on Applied Mathematics*, vol. 57.5, pp. 1328-1342, 1997.

[22] K. Amunts, C. Lepage, L. Borgeat, H. Mohlberg, T. Dickscheid, M.-É. Rousseau, S. Bludau, P.-L. Bazin, L. B. Lewis, A.-M. Oros-Peusquens, N. J. Shah, T. Lippert, K. Zilles and A. C. Evans, "BigBrain: An Ultrahigh-Resolution 3D Human Brain Model," *Science,* vol. 340.6139, pp. 1472-1475, 2013.